\documentclass[letterpaper, 10 pt, conference]{ieeeconf}  

\IEEEoverridecommandlockouts                              

\usepackage{mwe}
\usepackage{siunitx}
\usepackage[figurename=Fig.]{caption}
\usepackage[hyphens]{url}
\usepackage[T1]{fontenc}
\usepackage{subfig}
\usepackage{booktabs}
\usepackage{multirow}
\usepackage{amsmath}
\usepackage{hyperref}

\usepackage{tikz}

\newcommand\copyrighttext{
	\footnotesize 
	\textcopyright~2022 IEEE. Personal use of this material is permitted. Permission from IEEE must be obtained for all other uses, in any current or future media, including reprinting/republishing this material for advertising or promotional purposes, creating new collective works, for resale or redistribution to servers or lists, or reuse of any copyrighted component of this work in other works. DOI: \href{https://doi.org/10.1109/ITSC55140.2022.9922163}{10.1109/ITSC55140.2022.9922163}%
	}%
\newcommand\copyrightnotice{%
    \begin{tikzpicture}[remember picture,overlay]%
 	\node[anchor=south, xshift=0pt, yshift=4pt] at (current page.south)%
 	{\fbox{\parbox{\dimexpr\textwidth-\fboxsep-\fboxrule\relax}{\copyrighttext}}};%
 	\end{tikzpicture}%
}%

\title{\LARGE \bf Identification of Threat Regions From a Dynamic Occupancy Grid Map for Situation-Aware Environment Perception}

\author{Matti Henning$^1$, Jan Strohbeck$^1$, Michael Buchholz$^1$ and Klaus Dietmayer$^1$
\thanks{$^{1}$M. Henning, J. Strohbeck, M. Buchholz and K. Dietmayer are with the Institute of Measurement, Control and Microtechnology at Ulm University, 89081, Ulm, Germany. E-Mail: <firstname>.<lastname>@uni-ulm.de}%
\thanks{This research is accomplished within the UNICAR\emph{agil} project (FKZ
16EMO0290). We acknowledge the financial support for the project by the
German Federal Ministry of Education and Research (BMBF).}
}

\begin{document}

\maketitle
\thispagestyle{empty}
\pagestyle{empty}

\begin{abstract}
The advance towards higher levels of automation within the field of automated driving is accompanied by increasing requirements for the operational safety of vehicles.
Induced by the limitation of computational resources, trade-offs between the computational complexity of algorithms and their potential to ensure safe operation of automated vehicles are often encountered.
Situation-aware environment perception presents one promising example, where computational resources are distributed to regions within the perception area that are relevant for the task of the automated vehicle.
While prior map knowledge is often leveraged to identify relevant regions, in this work, we present a lightweight identification of safety-relevant regions that relies solely on online information.
We show that our approach enables safe vehicle operation in critical scenarios,  while retaining the benefits of non-uniformly distributed resources within the environment perception. 
\end{abstract}%

\section{Introduction}
\label{sec:intro}
\copyrightnotice%
Automated driving has received significant attention within the last decades. 
Both research and industrial applications try to achieve higher SAE levels of driving automation~\cite{SAE_2021}.
Generally, this is accompanied by increased hardware and software complexity, reflected in powerful computational units, more sensors, further diversity of sensor modalities, and multiple redundancies of sensor and hardware elements.
The reason for this increase is the corresponding increase in requirements for operational safety~\cite{Koopman2017, Riedmaier2020}, as responsibilities are shifted further towards the automated vehicle (AV) instead of its driver.

One central element for safe operation of AVs is environment perception. The increase in sensor hardware and corresponding algorithm complexity requires  significant amounts of computational resources and energy. 
One measure to reduce the required resources is \textit{situation-awareness}~\cite{Dahn2018}.
In situation-aware environment perception, the key idea is to distribute resources only towards regions that are identified as relevant for the current automation task of the vehicle. 
Thereby, compared to the naive \SI[mode=text]{360}{\degree} uniform distribution, the amount of data that needs to be processed, as well as the required computational resources, can be drastically reduced, which is ultimately reflected in the system's energy consumption.

For the identification of these regions, prior knowledge of the vehicle's environment is often used. Several approaches, e.g.,~\cite{nager2019, xu2015, homeier2011}, use an available high-definition map as a representation of relevant regions for environment perception. 
As traffic participants naturally operate on known areas of roads and sidewalks, this is an understandable approach for lower levels of automated driving.
However, two major concerns arise. First, available maps might be outdated, or temporary changes in the road topology are not considered. Second, with a rising level of automation, also unexpected behavior of traffic participants has to be considered.

\begin{figure}[!t]
    \centering
    \vspace{.45cm}
    \includegraphics[width=.24\linewidth]{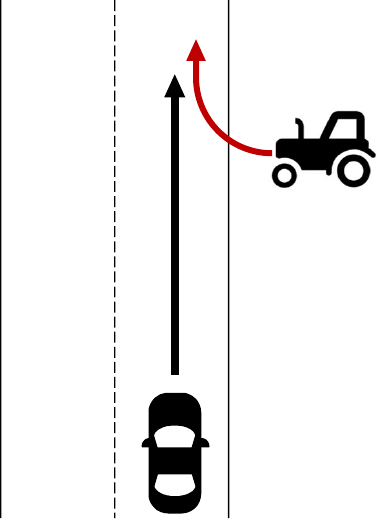}
    \caption{Scenario with risk-imposing traffic participant pulling out of an unmapped area into the road.}
    \label{fig:intro:collision}
\end{figure}

Hence, the assumption that all traffic participants solely operate on known, mapped areas becomes invalid. 
Examples might span from wildlife crossing rural roads to traffic participants pulling out of driveways or other unmapped road access points unexpectedly, as outlined in \figurename~\ref{fig:intro:collision} for a tractor pulling out of a field. 
Coherently, if the safe operation of automated driving shall be ensured, situation-aware environment perception needs to identify relevant regions not solely based on prior map knowledge. Instead, to consider the outlined shortcomings of map-based approaches, online information about the vehicle's environment, and especially its dynamic elements, is required to identify relevant regions.
In this work, we present our identification method based on online information and summarize our contribution as follows:
\begin{itemize}
    \item We present a lightweight identification of relevant regions corresponding to a potential collision threat, labeled \textit{threat regions}. For the identification of these regions, we leverage the model-free environment description of a dynamic occupancy grid map (DOGM)~\cite{Danescu2011} derived from lidar data.
    \item We evaluate our method in simulation, where we show that it enables safe operation of a situation-aware perception system by maintaining sufficient perceptive capabilities for the AV to act/react on. 
    \item Lastly, we verify the computational efficiency of the approach by evaluating the power consumption during post-processing of representative real-world data.
\end{itemize}

\section{Related Work}
\label{sc:rw}
The work presented in this paper relates to three research fields: situation-aware environment perception, threat assessment for automated driving, and object detection based on DOGMs. 

\subsection{Situation-Aware Environment Perception}
\label{sec:rw:saep}
As introduced in Section~\ref{sec:intro}, situation-awareness aims to distribute resources non-uniformly towards regions within the perceptive field that are deemed relevant for the current driving task~\cite{Bajcsy2018}.
Besides the outlined examples leveraging prior map information, a multitude of approaches to identify these regions exist, spanning from actuated sensor platforms, e.g., ~\cite{pellkofer2000, unterholzner2013}, to saliency-based attention mechanisms in deep-learning approaches~\cite{borji2019}.

Recently, we have presented a scalable and modular concept for situation-aware environment perception~\cite{henning2022}.
Our concept enables the integration of the above approaches and optimally configures the software modules of the environment perception processing chain based on requirements described in a multi-layer attention map (MLAM). 
The layers of the MLAM represent individual influences of the environment that define relevant regions for the respective situation of an AV. 
Prior map knowledge presents one example of such influences.
The requirement towards the environment perception results from the aggregation of all active attention layers, where the situation of the AV acts as an activation function for a subset of the available attention layers.

The work of this paper proposes one possible approach to identify relevant regions. 
By integrating the resulting description of threat regions into our recently presented concept of situation-aware environment perception as an attention layer, we enable the corresponding system to react dynamically on identified threat regions, while retaining the benefits of our concept.

\subsection{Threat Assessment for Automated Driving}
\label{sec:rw:threat}
Threat assessment is a central element for the safe operation of AVs. 
Next to the handling of internally safe operation, e.g., by ensuring stable vehicle dynamic controllers or resilience against hardware faults, threat assessment for automated driving represents the handling of external safe operation, i.e., it aims to identify the potential threat of known traffic participants interfering or engaging with the AV.
Within the familiar operation cycle of 1) sensing, 2) planning, and 3) acting, threats are usually identified in later stages of the sensing step, considered in the planning step and, coherently, reactive or proactive actions are executed. Li et al.~\cite{li2020} and Dahl et al.~\cite{dahl2018} have thoroughly reviewed the advances in threat assessment. They present a concise overview of employed methods, metrics, and applications.

The key element of threat assessment is the requirement for a reliable description of the traffic participant for which the threat shall be assessed.
Attention methods based on prior map knowledge restrict the perceptive capabilities and thereby suppress corresponding threat assessment outside the mapped area.
Within the scope of this work, our method for threat region identification leverages lightweight approaches for threat assessment at the beginning of the sensing step to enable the description of the traffic participant. 
Consequently, threat assessment of traffic participants within the identified threat regions is outside of the scope of this work.

\subsection{Object Detection Based on Occupancy Grid Maps}
A grid-based environment representation is a robust and efficient description of a vehicle's environment that divides the environment into cells and assigns, in principle, each cell a probability of it being occupied or free.
As the environment of AVs usually contains dynamic elements, e.g., other traffic participants, occupancy grid maps (OGMs) were quickly extended to estimate the velocity of each cell, resulting in dynamic occupancy grid maps (DOGMs).
A key benefit of the grid-based representation is that it requires neither spatial or movement models for objects nor associations of grid cells to objects. This enables an essentially unconstrained representation of an AV's environment.
~\cite{Danescu2011}. 

However, a model-based description of traffic participants is beneficial for interpreting and anticipating their behavior. This is especially valid considering the requirements for threat assessment (cf. Sec.\ref{sec:rw:threat}).
To extract model-based traffic participant representations from DOGMs, i.e., associating cell clusters of similar attributes to objects, existing approaches can be separated into cluster-based approaches and deep-learning approaches.
Cluster-based approaches, e.g., Steyer et al.~\cite{Steyer2017} or Gies et al.~\cite{Gies2018}, employ standard clustering algorithms like DBSCAN~\cite{ester1996}, to group occupied cells of similar attributes. They further plausibilize the resulting clusters by evaluation of the attributes of cells within the cluster or against attributes of their surrounding cells. 
Depending on the data fusion approach within the environment perception, clusters might be identified for the entire grid map (late fusion in~\cite{Gies2018}), or only new object hypotheses are generated from remaining cells not associated with existing object hypotheses (early fusion in~\cite{Steyer2017}). 
Deep-learning approaches, e.g.,~\cite{Piewak2017, Hoermann2018}, aim to overcome challenges of clustering approaches by intrinsically incorporating regional features within the employed network architectures to improve the resulting object detections. Approaches employing recurrent neural networks, e.g., Engel et al.~\cite{Engel2018}, incorporate the time domain to further stabilize estimates and reduce noise artifacts at increased complexity.  

The above approaches are commonly employed as input to model-based tracking algorithms. 
As the scope of our work lies on threat assessment in an early stage of the AV's operation cycle, we base our approach on lightweight, pre-filtered clustering algorithms. 

\section{System Architecture}
\label{sec:architecture}
\begin{figure}[!t]
    \centering
    \includegraphics[width=.95\linewidth]{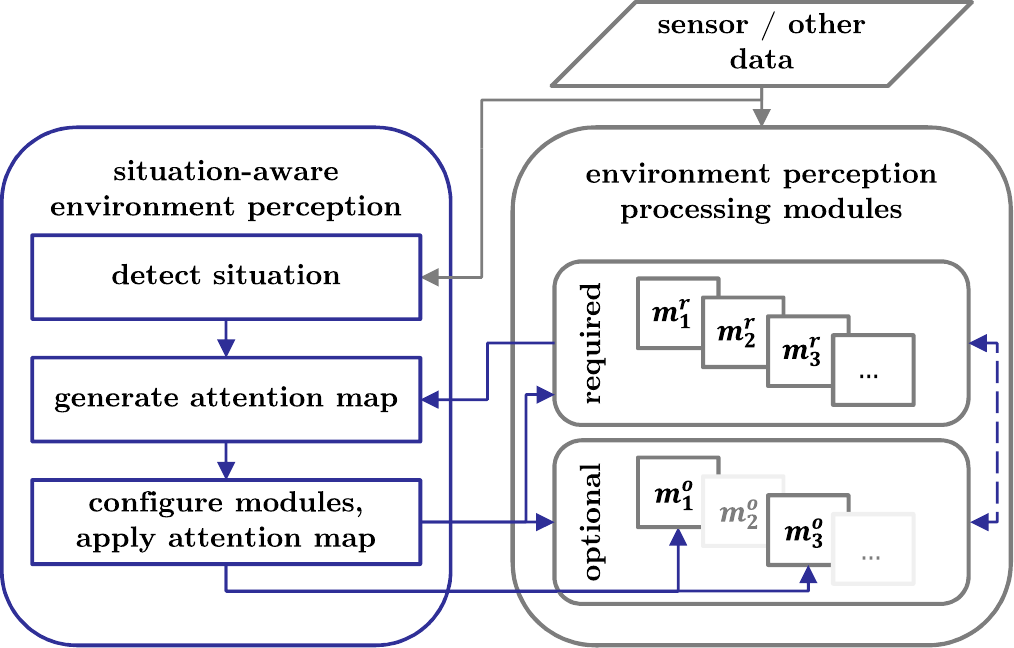}
    \caption{Interaction between situation-aware environment perception (blue) and an existing environment perception processing chain (gray), containing required software modules $m^r$ and optional modules $m^o$. Inactive modules are indicated by fading. Diagram adapted from~\cite{henning2022}.}
    \label{fig:method:architecture}
\end{figure}
In this section, we briefly outline the system architecture representing the interaction of our method for cluster-based threat region identification from DOGMs with our recently introduced concept of situation-aware environment perception, labeled \textit{awareness processing}~\cite{henning2022}. 
The system architecture is shown in Fig.~\ref{fig:method:architecture}. 

First, the situation is derived from sensor and other data, e.g., from external sources like infrastructure or other traffic participants. 
Second, based on the situation, corresponding attention layers are considered for the generation of the MLAM. Software modules used for the attention map generation, i.e., the processed output of a module is evaluated by an attention layer, are indicated as \textit{required}.
As the set of active attention layers depends on the situation, the sets of \textit{required} and \textit{optional} modules change dynamically. This is indicated by the dashed two-way arrow. 
Third, considering the requirements represented by the MLAM as well as the additional module requirements of the attention layer(s), the processing chain is configured. 
Lastly, the MLAM is applied intra-module-wise. Here, the non-uniform distribution of system resources towards relevant regions as per the MLAM is enforced. Modules required for the attention map generation remain unaffected to ensure the correct identification of new or changing relevant regions over time.

Our method for threat region identification (cf. Section~\ref{sec:method}) enables the introduction of an attention layer into the awareness processing that mitigates the challenges of map-based attention distribution.
Being based on a DOGM, modules required for its generation must be active when threat region identification is expected. 
As we focus the work of this paper on the method for threat region identification, integration and verification of our method in an application of awareness processing are outside of its scope.

\section{Threat Region Identification}
\label{sec:method}
This section presents our method for threat region identification in two steps.
First, clusters of non-stationary cells of a DOGM are identified.
We use the DOGM implementation presented by Nuss et al.~\cite{nuss2018} and generate clusters in an adapted approach from Gies et al.~\cite{Gies2018} using DBSCAN~\cite{ester1996}. 
Details of the used implementation and our adaptations are provided in Section~\ref{sec:method:identification}.
Second, the future occupied area of identified clusters is predicted, which is then evaluated for intersection with the AV's planned movement.

Clusters corresponding to an intersection are identified as threat regions. 
These regions are to be considered as requirements for perception modules, e.g., by representing them as an attention layer in the context of awareness processing (cf. Section.~\ref{sec:architecture}). 
As they do not represent a threat assessment of relevant traffic participants within the region, we emphasize that our method is designed to be efficient and straightforward. 
In this way, our method complies with the reduction in resource consumption pursued in situation-awareness, while enabling the description of traffic participants for threat assessment.

\subsection{Cluster Identification}
\label{sec:method:identification}
The DOGM presented by Nuss et al.~\cite{nuss2018} is generated from lidar data using a particle-based Dempster-Shafer probability hypothesis density (PHD)~\cite{mahler2003} / multi-instance Bernoulli~\cite{ristic2013} (MIB) filter. 
Each grid cell of size $a\times a$ corresponds to a state $s = \{ M_O, M_F, p, v, P_v\}$, 
comprising the belief masses for being occupied $M_O$ or free space $M_O$, the cell's 2D position $p$, as well as the cell's 2D velocity $v$ and covariance matrix $P_v$.
We refer to~\cite{nuss2018} for details.

The cluster identification is adapted from the methodology introduced by Gies et al.~\cite{Gies2018}, which is intended for object detection from the DOGM.
The approach first employs a search mask to neglect both unknown as well as unoccupied cells.
We define that stationary objects outside of known, driveable areas do not pose an external threat (cf. Section~\ref{sec:rw:threat}), so we further restrict the search mask to only include cells above a minimum required absolute velocity threshold.
Second, DBSCAN~\cite{ester1996} is used for the remaining cells to identify all clusters $\mathcal{C}$.
Every identified cluster
$c_k \in \mathcal{C}$ with $c_k = \{ s_j\}_{j \in \mathcal{J}_k}$ 
relates to its associated cells $j \in \mathcal{J}_k$ and their state vectors $s_j$.
Third, the identified clusters $\mathcal{C}$ are plausibilized w.r.t. occupancy probability, movement probability, and velocity variances of their associated cells in $\mathcal{J}_k$. 
In addition, we plausibilize the size of the cluster to neglect artifacts.

\subsection{Occupied Area Prediction and Intersection Evaluation}
The future occupied area for a plausibilized cluster is predicted over a prediction horizon $T$ using its orientation $\varphi_k$, position $p_k$, absolute velocity $\bar{v}_k$, and object box estimate $\boldsymbol{B}_k$. 
The position is estimated by averaging the corresponding state elements of associated cells. 
The cluster orientation and absolute velocity are derived from their 2D velocity $v$.
Lastly, $\boldsymbol{B}_k$ is derived from the positions of all contained cells and oriented alongside $\varphi_k$.
The occupied area prediction $\text{pred}_{c_k}$ is then described by the convex hull that is spanned between the cluster's object box estimate $\boldsymbol{B}_k$ and its prediction at $T$ using a constant velocity (CV) model for the cluster's orientation $\varphi_k$ and velocity $\bar{v}_k$. 
The convex hull of an occupied area prediction can effectively be described by a set of points:
\begin{align}
    \label{eq:method:boundingbox}
    \text{pred} = \left\{ p_{\text{hull}_1}, p_{\text{hull}_2}, ..., p_{\text{hull}_n}\right\}.
\end{align}

Our method to estimate the future occupied area of a cluster within the prediction horizon relates to the group of kinematic-based threat assessment (cf.~\cite{li2020, dahl2018}). Similar to other approaches of its type, it is efficient to implement and easy to extend if corner cases deem that necessary.

To identify a cluster as a threat region, the intersection of its predicted occupied area with the planned movement of the AV $\text{pred}_{AV}$ is evaluated.
As the dimension and kinematics of the AV as well as its planned trajectory within the prediction horizon $T$ are known, $\text{pred}_{AV}$ can be represented as a set of points describing the closed hull as per \eqref{eq:method:boundingbox}.
With $\text{pred}_{AV}$ and $\text{pred}_{c_k}$ represented as a set of 2D points, the two areas intersect if at least one pair of line segments between two adjacent hull points of the AV and the cluster $c_k$ intersect. 
Consequently, 2D line segment intersection is evaluated for all combinations of line segments of $\text{pred}_{AV}$ and $\text{pred}_{c_k}$.

\begin{figure}[!t]
    \centering
    \includegraphics[width=.5\linewidth]{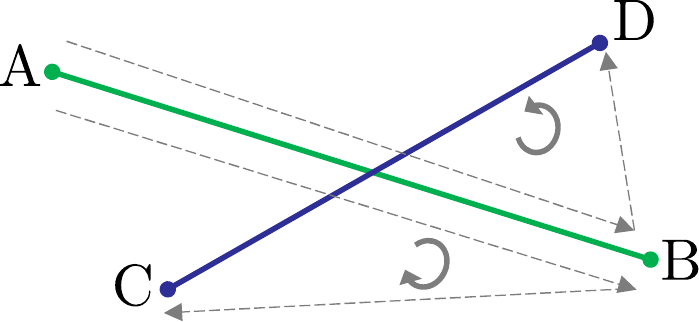}
    \caption{Example for two line segments that intersect. The orientation is indicated by a curved arrow for ordered point sequences $\overline{ABC}$ and $\overline{ABD}$ that are indicated by dashed arrows. Figure adapted from~\cite{cormen2009}.}
    \label{fig:method:intersection}
\end{figure}
While many methods for 2D line segment intersection exist, we employ the method of Cormen et al.~\cite{cormen2009}. 
They evaluate the orientation of combinations of three ordered points from the four points describing the two line segments. 
For the example shown in Fig.~\ref{fig:method:intersection}, the segments $\overline{AB}$ and $\overline{CD}$ intersect, as both $\overline{ABC}$ and $\overline{ABD}$, as well as $\overline{CDA}$ and $\overline{CDB}$ have different orientations.
Although the consideration of collinearity is possible, we evaluate only true line segment intersection here. The case that one pair of line segments is collinear and all other pairs do not intersect can only occur where the cluster's future occupied area is parallel to the AV's prediction. 
In this case, the cluster is either already on the AV's planned trajectory, or it does not pose a threat in terms of collision risk. 
Similarly, if $\text{pred}_{c_k}$ is contained within $\text{pred}_{AV}$, the cluster is on the AV's planned trajectory and no additional threat region identification is required.

Finally, the identified threat regions, i.e., clusters that relate to any intersection of line segments between $\text{pred}_{c_k}$ and $\text{pred}_{AV}$, need to be considered in the environment perception of an AV so that the potential collision threat of traffic participants within the identified regions can be assessed. 
We have outlined the integration as an additional attention layer in situation-aware environment perception in Section~\ref{sec:architecture}.

\section{Evaluation}
\label{sec:eval}
Closed-loop testing of safety-critical features in real-world conditions is both dangerous as well as cost-intensive. 
Therefore, we evaluate our introduced method for identifying threat regions using DeepSIL~\cite{Strohbeck2021}, a recently published closed-loop simulation framework. 
The framework can produce trajectories of vehicles alongside a pre-defined map using the lanelet2~\cite{poggenhans2018} format, which can be easily generated. 
Further, it provides ground-truth kinematic data and a simulated sensor for generating a DOGM. 
Instead of the resource-intensive multiple trajectory prediction network (MTP), the vehicle trajectories for our evaluation are generated from the efficient intelligent driver model (IDM)~\cite{treiber2000}, which is used as a baseline in~\cite{Strohbeck2021}.

The intrinsic threat imposed by a traffic participant is often related to time-based metrics (cf.~\cite{li2020, dahl2018}), e.g., the time to collision (TTC) or the time to react (TTR). 
Our work focuses on enabling early detection of potentially dangerous traffic participants without considering the intrinsic kinematic relations. 
Therefore, to evaluate the effectiveness of our method, we define the TTR to reflect the duration between the time when its detection becomes possible (time of earliest detection, ToD) and the time of collision (ToC).
The benefit of our approach is then reflected in the relative increase of the time to react (riTTR)
\begin{equation}
    \text{riTTR} = \frac{\text{ToD}_{\text{ours}}-\text{ToC}}{\text{ToD}_{\text{prior}}-\text{ToC}}-1,
\end{equation}
which is represented by the relation between our method and a detection method solely based on prior map knowledge. 
We determine $\text{ToD}_\text{ours}$ as the first time of intersection of $\text{pred}_{AV}$ and $\text{pred}_{c_k}$. 
Further, we determine $\text{ToD}_\text{prior}$ as the first time any point of $\boldsymbol{B}_k$ intersects with $\text{pred}_{AV}$, representing the time a traffic participant has sufficiently entered a perception area solely based on prior map knowledge.

We have outlined plausible real-world examples for the necessity of online threat region identification in Section~\ref{sec:intro}. 
To identify reasonable scenarios for evaluating our method, we refer to established safety assessment protocol descriptions, e.g., by NHTSA~\cite{Rao2021}, that are designed for automated emergency braking (AEB). The scenario descriptions can be separated into three groups: approaching and lane changing on a straight path, straight crossing paths, and turning towards or away from the AV's orientation. 
While most scenarios are covered with the assumption of prior map knowledge, the most challenging scenarios refer to the group of turning traffic participants, as indicated in Fig.~\ref{fig:intro:collision}.
Consequently, within the scope of this work, we evaluate the riTTR in two turning constellations that reflect challenging, risk-imposing scenarios.

\begin{figure}[t!]
    \centering
    \subfloat[Traffic participant turning into the AV's path.]{
    \includegraphics[width=\linewidth]{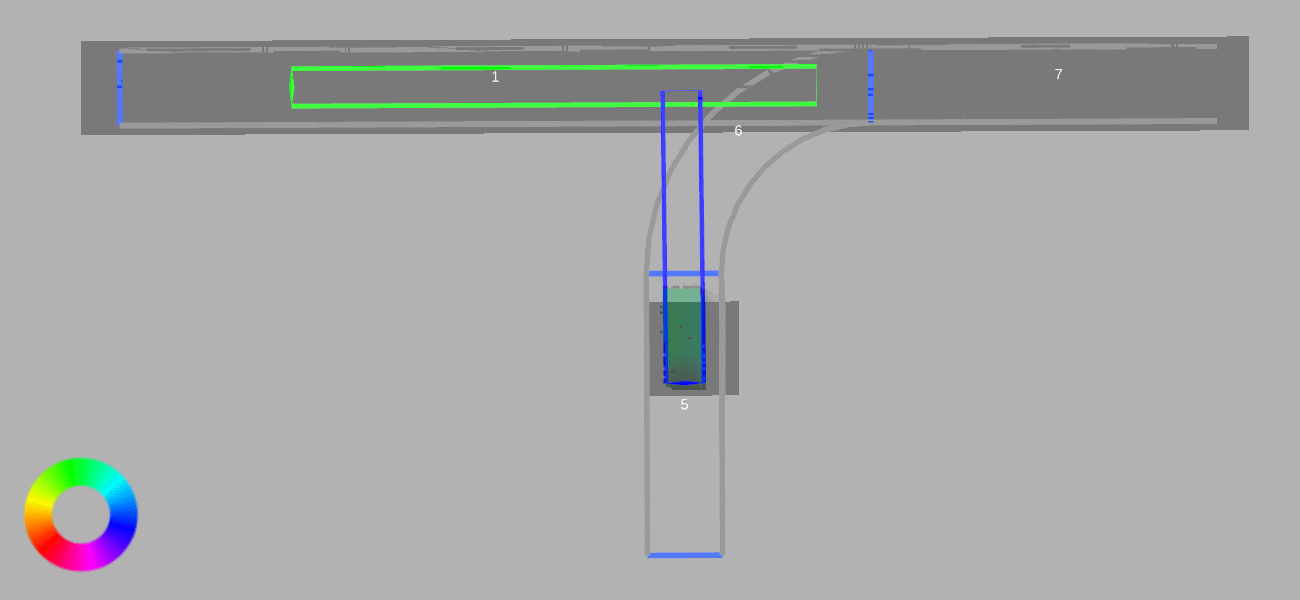}
    \label{fig:eval:turnin}}
    \hfil
    \subfloat[Traffic participant turning over the AV's path.]{
        \includegraphics[width=\linewidth]{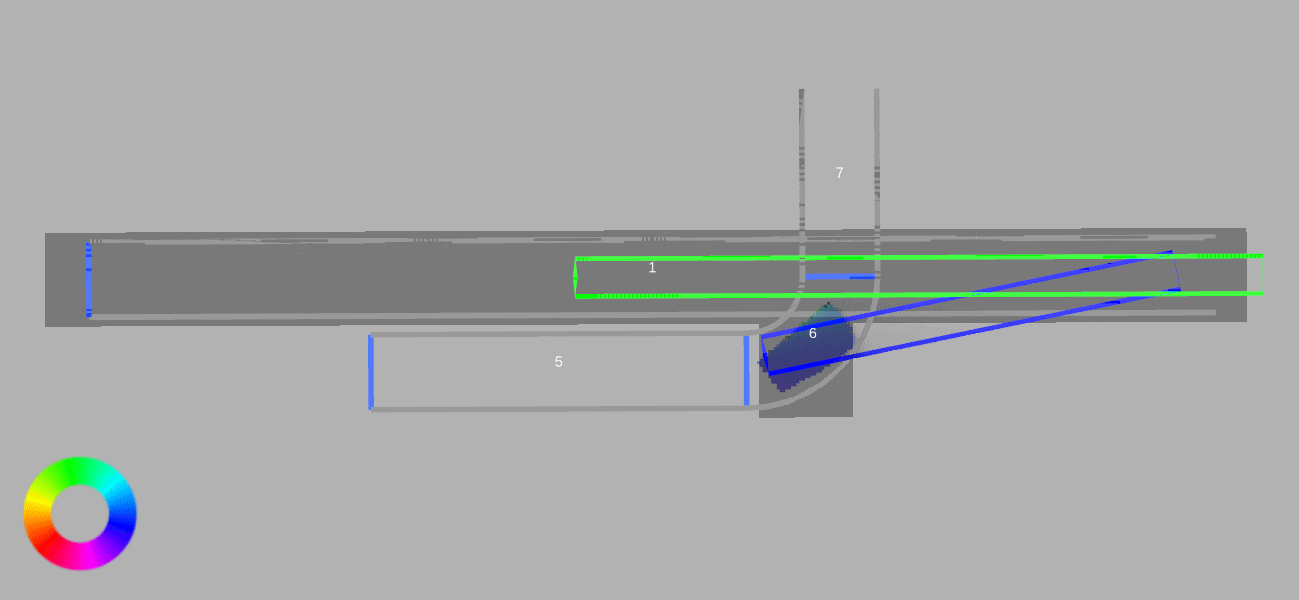}
        \label{fig:eval:turnover}}
    \hfil
    \caption{Simulation results for $T=3\si{\second}$ at $\text{ToD}_{\text{ours}}$. $\text{pred}_{AV}$ and $\text{pred}_{c_k}$ are indicated in green and blue, respectively. DOGM cell orientation is color-coded as per the color circle. The derived perception area is indicated by dark-gray rectangles.}
    \label{fig:eval:scenarios}
\end{figure}
The simulation results for $T=3~\si{\second}$, reflecting a commonly chosen prediction horizon, are visualized in Fig.~\ref{fig:eval:scenarios} at $\text{ToD}_{\text{ours}}$. The figure shows $\text{pred}_{AV}$ in green and $\text{pred}_{c_k}$ in blue.
The orientation of DOGM cells is color-coded as indicated by the color-code circle. Further, the lane boundaries are outlined, where only the straight lane is contained in the prior map knowledge. 
The dark-gray rectangles describe the derived perception area as per situation-aware environment perception (cf. Section~\ref{sec:rw:saep}). 
Consequently, such an area at a cluster's position reflects its identification as a threat region.
The numerical results are summarized in Tab.~\ref{tab:eval:riTTR} and will be discussed in the following.

A turning in scenario is shown in Fig.~\ref{fig:eval:turnin}, representing the scenario described in Fig.~\ref{fig:intro:collision}. 
The initial movement of the traffic participant is similar to a straight crossing scenario so that the threat is well-modeled by the CV model for $\text{pred}_{c_k}$. 
Consequently, $\text{ToD}_{\text{ours}}$ increases to \SI[mode=text]{2.1}{\second}. 
A riTTR of \SI[mode=text]{425}{\percent} enables a safe operation of the AV.

Fig.~\ref{fig:eval:turnover} shows a turning over scenario, where a traffic participant first moves parallel to the AV and then turns into its path.
The riTTR of \SI[mode=text]{60}{\percent}, reflecting an increase from $\text{ToD}_{\text{prior}}=\SI[mode=text]{0.5}{\second}$ to $\text{ToD}_{\text{ours}}=\SI[mode=text]{0.8}{\second}$, is still significant and an improvement towards safe operation can be assumed.
Further, Fig.~\ref{fig:eval:turnover} shows that the estimated cluster orientation does not match the actual orientation of the simulated traffic participant. 
This is a known effect of the used implementation of the DOGM (cf. Nuss et al.~\cite{nuss2018}).
For extended objects, the CV particle model is unable to represent a turning object properly. 
Instead, a predicted particle of one position of the object is confirmed at another position of the object. A change in orientation is reflected at first only at the respective object edges. 
As the cluster orientation $\varphi_k$ is derived from cell averages, it consequently diverges from the actual object orientation.

With respect to our introduced CV model for deriving $\text{pred}_{c_k}$, a straightforward adaptation to mitigate this effect is the introduction of a cluster angle uncertainty $\varphi_u$, with $\tilde{\varphi_k} = \varphi_k + \varphi_u$,
so that the rectangular shape of $\text{pred}_{c_k}$ is transformed into a symmetrical trapezoid shape. 
Introducing an uncertainty of $\varphi_u = 10^{\circ}$ for the turning over scenario, $\text{ToD}_{\text{ours}}$ increases to \SI[mode=text]{1.1}{\second} and riTTR increases to \SI[mode=text]{120}{\percent} respectively. 
\begin{table}[t!]
    \centering
    \caption{Numerical results for the scenarios in Fig.~\ref{fig:eval:scenarios}. Two cluster angle uncertainties of \SI[mode=text]{0}{\degree} and \SI[mode=text]{10}{\degree} are compared.}
    \label{tab:eval:riTTR}
    \setlength{\tabcolsep}{9pt}
    \begin{tabular}{r c  c c  c c}
    \toprule
        \multirow{2}{*}{scenario}   & \multirow{2}{*}{prior} & \multicolumn{2}{c}{$\varphi_u=0^\circ$} & \multicolumn{2}{c}{$\varphi_u=10^\circ$}   \\
                                    &  & ours &  riTTR & ours &  riTTR\\
       \midrule
    turning in                  & \SI[mode=text]{0.4}{\second} & \SI[mode=text]{2.1}{\second}& \SI[mode=text]{425}{\percent} & \SI[mode=text]{2.1}{\second} & \SI[mode=text]{425}{\percent}\\
    turning over                & \SI[mode=text]{0.5}{\second} & \SI[mode=text]{0.8}{\second} & \SI[mode=text]{60}{\percent}  & \SI[mode=text]{1.1}{\second} & \SI[mode=text]{120}{\percent}\\
    \bottomrule 
    \end{tabular}
\end{table}
An alternative to model adaptations is provided by Schreiber et al.~\cite{Schreiber2021}. They present a promising deep-learning-based approach for generating DOGMs using a recurrent network architecture. Their work is motivated explicitly by the indicated shortcomings of the CV particle filter model. However, having the conciseness of this paper and the already good results of our model adaptation in mind, further evaluation of mitigation methods is omitted.

\section{System Power Consumption}
\begin{figure}[!t]
    \centering
    \includegraphics[width=\linewidth]{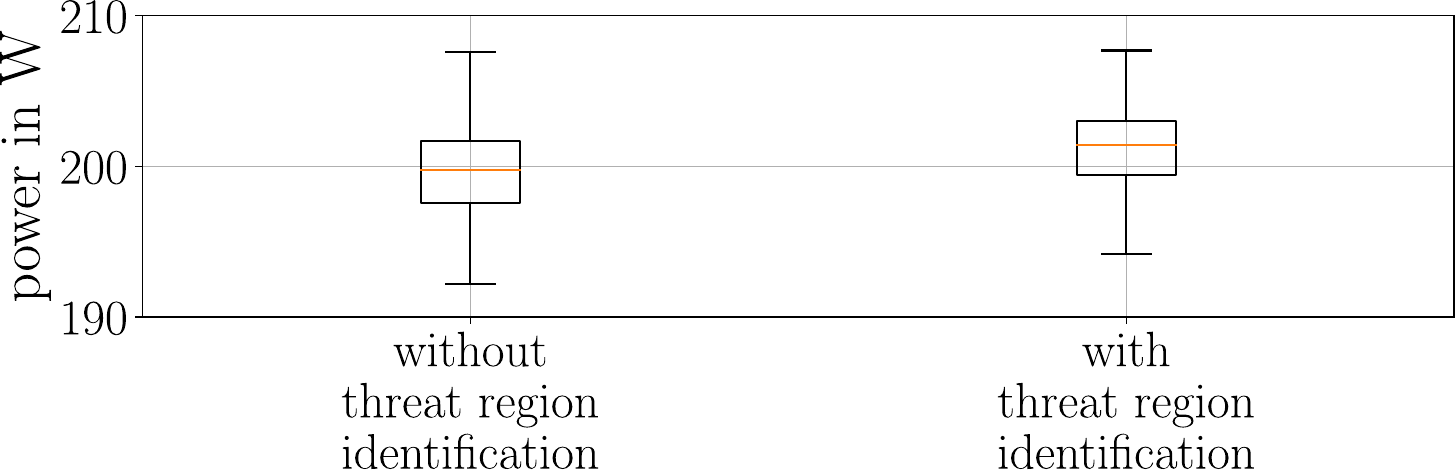}
    \caption{System power consumption comparison between baseline and included threat region identification. Whiskers correspond to 3$\times$inter-quartile range.}
    \label{fig:power:comparison}
\end{figure}
In the previous section, we verified that our method enables safe operation of an AV in risk-imposing scenarios.
In this section, we evaluate the system power consumption as a measure for added complexity since computational resources aboard an AV are very limited, and their usage shall be minimized within the context of situation-aware environment perception.
We added our method of threat region identification to our perception processing chain of an AV (cf.~\cite{Gies2018}) and post-processed sensor data from a representative route within the vicinity of Ulm University. 
Post-processing is conducted on a high-level consumer-grade PC using an AMD Ryzen Threadripper 2990WX CPU, two NVIDIA GeForce RTX 2080 Ti GPUs, and 64GB of RAM.
The route comprises approximately \SI[mode=text]{5}{\km} with various aspects of urban and rural driving as well as a short highway-like section. 
It does not contain any scenarios corresponding to an actual threat of collision.
The data were captured with our automated test vehicle, equipped with a range of sensors, see \cite{Buchholz2021} for details.

The system power consumption, measured during the post-processing of the route, is averaged from 100 samples per second and shown in Fig.~\ref{fig:power:comparison} as boxplots.
The figure compares the power consumption distribution between the baseline, containing only processing modules required to generate our particle-based DOGM and the additional threat region identification module.
The results verify that our approach is lightweight, as the median power consumption is increased by only \SI[mode=text]{0.8}{\percent}.
Hence, our method presents a viable approach for the identification of threat regions for situation-aware environment perception.

\section{Conclusions}
In this work, we have presented a lightweight identification of \textit{threat regions} corresponding to a potential collision risk of traffic participants with the automated vehicle.
For the identification of these regions, our method requires no prior knowledge of the environment but uses a clustering approach for cells of a dynamic occupancy grid map. This is combined with a model-based prediction to evaluate the intersection of a cluster's  future occupied area with the AV's known trajectory.
Contrary to approaches relying solely on prior map knowledge for situation-aware environment perception, we have shown that our prior-independent approach significantly increases the reaction time for an AV in the evaluated collision scenarios, while increasing the median system power consumption by only \SI[mode=text]{0.8}{\percent}.
By integrating our method for threat region identification as an attention layer into our previously presented concept for situation-aware environment perception (cf.~\cite{henning2022}), we enable a safe operation of AVs in terms of perceptive capabilities, while maintaining the benefits of situation-awareness. 
Further, we have outlined countermeasures to resolve challenges related to the underlying constant velocity (CV) model for particles of the chosen DOGM implementation.
In our continued research, we will pursue the outlined countermeasures, e.g., model adaptations or other derivations of the DOGM like from \cite{Schreiber2021}. 


\end{document}